\documentclass{article}
\usepackage{spconf,amsmath,graphicx}
\usepackage{epstopdf,booktabs,multicol,multirow,enumitem,footnote,type1cm,amssymb,calc}
\usepackage{color,xcolor}
\usepackage[linesnumbered,ruled,vlined]{algorithm2e}
\SetKwInput{KwInit}{Initialization}                % Set the Input
\SetKwInput{KwOutput}{Output}              % set the Output
\SetCommentSty{mycommfont}
\title{U2++ MoE: Scaling 4.7x Parameters with minimal Impact on RTF}
\name{
    \begin{tabular}{c}
    Xingchen Song$^{1,2\dagger}$\thanks{$\dagger$ Corresponding Author.},~Di Wu$^{1,2}$,~Binbin Zhang$^{1,2}$,\\~Dinghao Zhou$^{2}$,~Zhendong Peng$^{2}$,~Bo Dang$^{2}$,~Fuping Pan$^{1}$,~Chao Yang$^{1,2}$
    \end{tabular}
}
\address{
    $^1$GuaSemi Inc., Beijing, China~~
    $^2$WeNet Open Source Community \\
    xingchen.song@gua.com$^{\dagger}$
    \vspace{-10pt}
}

\begin{document}

\maketitle

\begin{abstract}

% 0. simplicity
% 1. streaming
% 2. rtf analyze

Scale has opened new frontiers in natural language processing, but at a high cost. In response, by learning to only activate a subset of parameters in training and inference, Mixture-of-Experts (MoE) \cite{DBLP:conf/iclr/ShazeerMMDLHD17,DBLP:conf/iclr/LepikhinLXCFHKS21} have been proposed as an energy efficient path to even larger and more capable language models and this shift towards a new generation of foundation models is gaining momentum, particularly within the field of Automatic Speech Recognition (ASR). Recent works \cite{DBLP:journals/corr/abs-2305-15663,DBLP:conf/interspeech/YouFSY21,DBLP:journals/corr/abs-2307-05956,DBLP:conf/iscslp/YouFSY22} that incorporating MoE into ASR models have complex designs such as routing frames via supplementary embedding network, improving multilingual ability for the experts, and utilizing dedicated auxiliary losses for either expert load balancing or specific language handling. We found that delicate designs are not necessary, while an embarrassingly simple substitution of MoE layers for all Feed-Forward Network (FFN) layers is competent for the ASR task. To be more specific, we benchmark our proposed model on a large scale inner-source dataset (160k hours), the results show that we can scale our baseline Conformer (Dense-225M) to its MoE counterparts (MoE-1B) and achieve Dense-1B level Word Error Rate (WER) while maintaining a Dense-225M level Real Time Factor (RTF).
% Besides, to the best of our knowledge, we are the first to show that MoE works well on streaming mode,
Furthermore, by applying Unified 2-pass framework with bidirectional attention decoders (U2++) \cite{DBLP:journals/corr/abs-2203-15455}, we achieve the streaming and non-streaming decoding modes in a single MoE based model, which we call U2++ MoE.
We hope that our study can facilitate the research on scaling speech foundation models without sacrificing deployment efficiency.

\end{abstract}


\section{Introduction}
\label{sec:intro}

Scaling up neural network models has recently received great attention, given the significant quality improvements on a variety of tasks including natural language processing \cite{radford2019language,DBLP:conf/nips/BrownMRSKDNSSAA20} and speech processing \cite{DBLP:conf/icml/RadfordKXBMS23,DBLP:journals/corr/abs-2303-01037}.

While training massive models on large amounts of data can almost guarantee improved quality, there are two factors affecting their practicality and applicability: (1) \textit{training efficiency} and (2) \textit{inference efficiency}. Large dense models are often prohibitively compute-intensive to train, with some models requiring TFlops-days of compute \cite{DBLP:conf/nips/BrownMRSKDNSSAA20,DBLP:conf/emnlp/KuduguntaHBKLLF21}. A recent line of work has proposed sparsely-gated Mixture-of-Experts (MoE) layers \cite{DBLP:conf/iclr/ShazeerMMDLHD17,DBLP:conf/iclr/LepikhinLXCFHKS21} as an efficient alternative to dense models in order to address both training and inference efficiency limitations.

There have been several related Mixture-of-Expert approaches for ASR modeling \cite{DBLP:journals/corr/abs-2305-15663,DBLP:conf/interspeech/YouFSY21,DBLP:journals/corr/abs-2307-05956,DBLP:conf/iscslp/YouFSY22}. In those models each frame of the input sequence activates a different subset of the experts, hence the computation cost per frame becomes only proportional to the size of the activated sub-network. To avoid collapse to just a few experts while ignoring all others, all of those works use load balancing mechanisms such as dedicated auxiliary losses \cite{DBLP:conf/iclr/LepikhinLXCFHKS21,DBLP:journals/jmlr/FedusZS22}. Nonetheless, the resulting complex optimization objectives often lead to a large amount of hyper parameter tuning, such as the weight of each auxiliary loss. Moreover, load balancing is designed to address the issue of expert sparsity in the NLP field when routing different tokens. However, this issue may not hold in the speech domain, as there is a high degree of similarity between neighboring speech frames \cite{is19/apc}. Forcing speech frames to be evenly distributed among all experts does not align with intuition, as it conflicts with the natural continuity observed in the relationships between adjacent speech frames.

\begin{figure*}[!ht]
    \centering
    \includegraphics[scale=0.09]{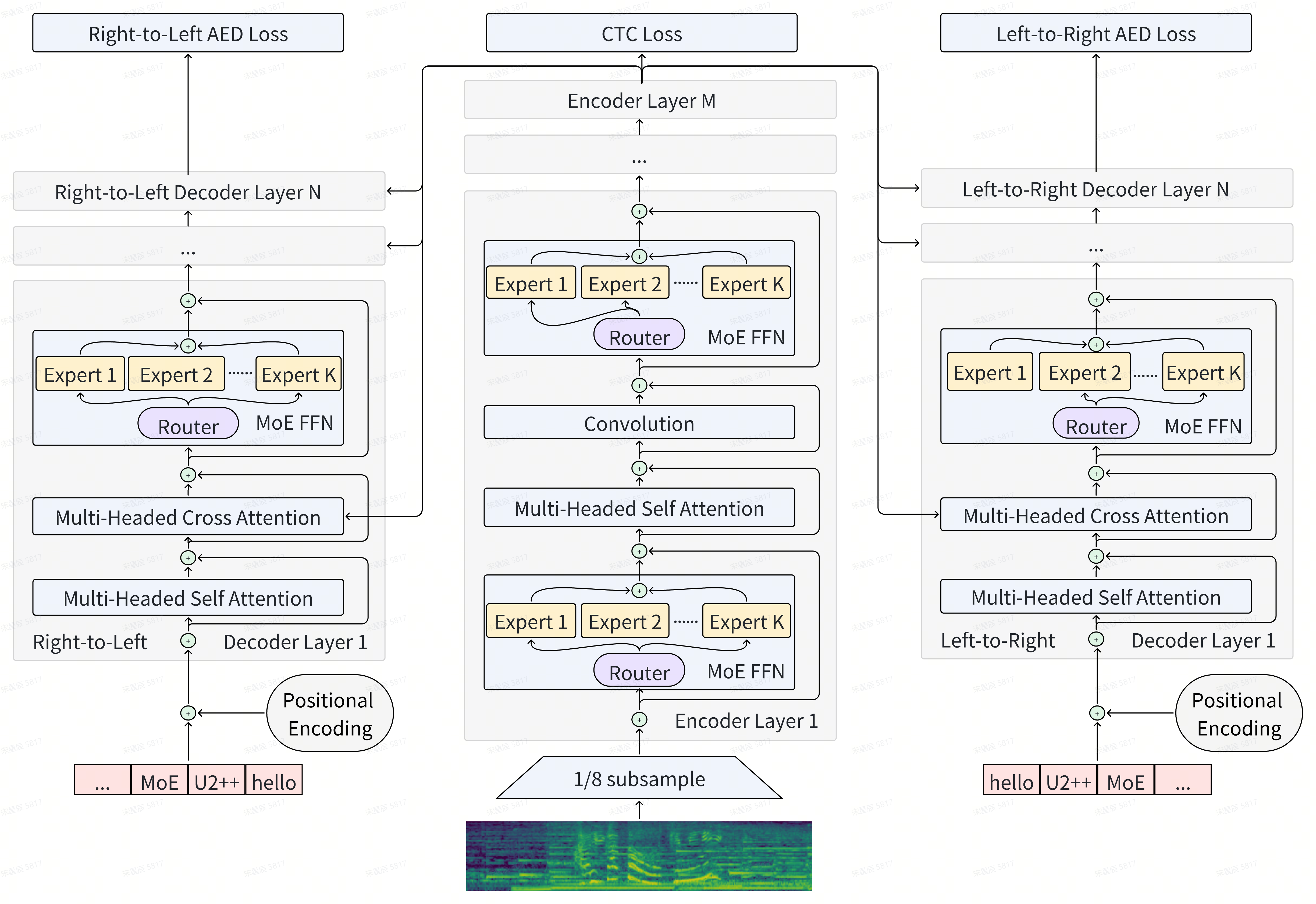}
    \caption{The proposed U2++ MoE, a unified (streaming and non-streaming) two-pass (encoder for 1st pass decoding and decoder for 2nd pass rescoring) joint CTC/AED framework, enhanced with bidirectional decoders and Mixture-of-Experts. For efficient compression of speech frames, we employ 1/8 subsampling and structure our architecture with $M$ encoder layers alongside $2N$ decoder layers, wherein equal divisions of $N$ layers are allocated to both the right-to-left and left-to-right decoders.}
    \vspace{-10pt}
    \label{fig:u2pp-moe}
\end{figure*}

Despite several notable successes of speech MoE, wide-spread adoption has been hindered by training complexity and the lack of streaming capabilities. We address these with the introduction of U2++ MoE. We simplify the integration of MoE and preclude the necessity for any auxiliary losses. Our proposed method mitigate the complexities, and we show large sparse models may be trained, for the first time, with unified streaming \& non-streaming fashion.

\section{Related Works}

Several mixture-of-expert strategies have been developed for enhancing ASR modeling, but our work differs from them in the following ways.

\begin{itemize}
    \item[1)] In contrast to all prior studies \cite{DBLP:journals/corr/abs-2305-15663,DBLP:conf/interspeech/YouFSY21,DBLP:journals/corr/abs-2307-05956,DBLP:conf/iscslp/YouFSY22}, our MoE model do not include any auxiliary losses for expert routing, thus significantly streamlining the training optimization process.
    \item[2)] Compared to \cite{DBLP:conf/interspeech/YouFSY21,DBLP:journals/corr/abs-2307-05956,DBLP:conf/iscslp/YouFSY22}, our MoE study also works without using any shared embedding networks, thereby simplifying the model architecture and enhancing its generality for model scaling.
    \item[3)] Compared to all previous works \cite{DBLP:journals/corr/abs-2305-15663,DBLP:conf/interspeech/YouFSY21,DBLP:journals/corr/abs-2307-05956,DBLP:conf/iscslp/YouFSY22}  that exclusively explored the application of MoE layers within the encoder, our study extends this innovation by integrating MoE layers into the decoder's FFN as well. Notably, \textit{You et al.} \cite{DBLP:conf/iscslp/YouFSY22} have also attempted to modify all FFN modules in encoder into MoE layers, but it fails to achieve a better performance (\textit{detailed in \cite{DBLP:conf/iscslp/YouFSY22}, section 3.2, paragraph 1, last sentence}). In contrast, we are the first to demonstrate the effectiveness of MoE layer substitution across both encoder and decoder components.
    \item[4)] We are the pioneers in demonstrating the streaming capability of the MoE. While \textit{Hu et al}. \cite{DBLP:journals/corr/abs-2305-15663} have made attempts to integrate MoE layers into a causal encoder to enable streaming recognition, their approach resulted in a notable deterioration in the average WER (\textit{detailed in \cite{DBLP:journals/corr/abs-2305-15663}, section 5.1.1, paragraph 2, first sentence}). In stark contrast, our approach, which marries the MoE-based Conformer with the U2++ framework, successfully facilitates both streaming and non-streaming decoding modes within a singular MoE-based model.
    \item[5)] Our research primarily emphasizes scaling models without a notable increase on RTF, diverging from prior efforts that predominantly concentrate on enhancing the accuracy of multi-lingual or multi-accent recognition \cite{DBLP:journals/corr/abs-2305-15663,DBLP:conf/interspeech/YouFSY21,DBLP:journals/corr/abs-2307-05956,DBLP:conf/iscslp/YouFSY22}. These studies lack a comprehensive analysis of inference latency, such as Dense-1B model v.s. MoE-1B model or Dense-225M model v.s. MoE-1B model. In this paper, however, we demonstrate that a MoE-1B model can achieve the accuracy of a Dense-1B model while maintaining the inference efficiency of a Dense-225M model.
\end{itemize}

In summary, our guiding principle has been to \textit{\textbf{keeping MoE model as simple as possible and is thus more generic for scaling up models}}. Our model do not require any auxiliary losses or any additional embedding networks. By applying 1) an embarrassingly simple replacement of all FFN layers with MoE layers and 2) the U2++ framework to Conformer \cite{DBLP:conf/interspeech/GulatiQCPZYHWZW20}, we prove that MoE-1B model can achieve Dense-1B level accuracy with Dense-225M level inference cost, alongside the capability for streaming.

\section{Methodology}
\label{sec:method}

Our model uses Conformer (for encoders) and Transformer (for decoders) as the main building block. A Conformer encoder layer \cite{DBLP:conf/interspeech/GulatiQCPZYHWZW20} consists of a multi-headed self-attention and a convolution-based layer sandwiched by two FFN. A Transformer decoder layer \cite{nips17/transformer} consists of a multi-headed self-attention, a multi-headed src-attention and one FFN. As shown in Fig.\ref{fig:u2pp-moe}, to incorporate experts, we use an MoE layer \cite{DBLP:conf/iclr/ShazeerMMDLHD17,DBLP:conf/iclr/LepikhinLXCFHKS21} to replace all FFN in the encoders and decoders. Similar to \cite{DBLP:conf/iclr/ShazeerMMDLHD17,DBLP:conf/iclr/LepikhinLXCFHKS21}, the MoE layer consists of a routing network and multiple experts, each of which is an FFN.

We use the joint Connectionist Temporal Classification (CTC) loss \cite{icml06/ctc} and Autoregressive Encoder Decoder (AED) loss \cite{nips17/transformer} for training the proposed model. The combined loss has two hyper parameters ($\lambda$ and $\alpha$) to balance the importance of different losses (\textit{more details can be found in \cite{DBLP:journals/corr/abs-2203-15455}, section 2.1}):
\begin{equation}
    L = \lambda L_{CTC} + (1 - \lambda)(\alpha L_{AED}^{right2left} + (1 - \alpha) L_{AED}^{left2right})
\end{equation}

Similar to U2 \cite{DBLP:conf/interspeech/YaoWWZYYPCXL21}, we adopt the dynamic chunk masking strategy to unify the streaming and non-streaming modes. Firstly, the input is split into several chunks by a fixed chunk size $C$ and every chunk attends on itself and all the previous chunks, so the whole latency for the CTC decoding in the first pass only depends on the chunk size. When the chunk size is limited, it works in a streaming way; otherwise it works in a non-streaming way. Secondly, the chunk size is varied dynamically from 1 to the max length of the current training utterance in the training, so the trained model learns to predict with arbitrary chunk size.

\section{Experiments}

\subsection{Datasets}
Our training corpus comprises mixed datasets gathered from a variety of application domains, amounting to a substantial 160k hours of large-scale, industrial-level training data. This corpus consists predominantly of Mandarin (90\%) with the remainder in English (10\%).

To evaluate the capabilities of the proposed method, we use the most widely used benchmark for the Mandarin ASR task, namely SpeechIO TIOBE ASR Benchmark \footnote{https://github.com/SpeechColab/Leaderboard}. SpeechIO test sets are carefully curated by SpeechIO authors, crawled from publicly available sources (Youtube, TV programs, Podcast etc), covering various well-known scenarios and topics (TV News, VLog, Documentary and so on), transcribed by payed professional annotators thus is exceptionally suitable for testing a model's general speech recognition capabilities. Cumulatively, the 26 publicly available SpeechIO test sets amount to 60.2 hours, averaging 2.3 hours of data across each domain.

\subsection{Training Details}
\label{sec:train_detail}
In all experiments, we utilize 80-dimensional log-mel filterbank features, computed using a 25ms window that is shifted every 10ms. Each frame undergoes global mean and variance normalization. For modeling Mandarin, we employ character-based representations, whereas for English, we utilize byte-pair encoding (BPE), culminating in a comprehensive vocabulary of 6000 units. All our experiments are conducted in WeNet toolkit \cite{DBLP:journals/corr/abs-2203-15455} with DeepSpeed \cite{DBLP:conf/kdd/RasleyRRH20} enabled, all the models are trained using 8 * NVIDIA 3090 (24GB) GPUs.

We have developed three distinct models, as detailed in Table \ref{tab:model_arch}, all of which adopt the parameters $Head = 8$, $CNN_{kernel} = 15$, $\lambda = 0.3$, and $\alpha = 0.3$. In the context of the MoE layer, we configure it with 8 experts and enable only the top two experts during both the training and inference phases. For the decoding process, the CTC decoder initially generates the N-Best hypotheses during the first pass. Subsequently, these hypotheses are rescored by the attention decoder in the second pass to produce the final outcomes.

\vspace{-10pt}
\begin{table}[!ht]
	\centering
	\caption{Configuration of different models.}
	\scalebox{0.9}{
		\begin{tabular}{c|cccc}
			\toprule[1.5pt]
			(a) Model & (b) $M$ & (c) $N$ & (d) $d^{ff}$ & (e) $d^{att}$\\
                \midrule
                Dense-225M & 12 & 3 & 2880 & 720\\
                Dense-1B & 32 & 6 & 4096 & 1024 \\
                MoE-1B & 12 & 3 & 2880 & 720 \\
			\bottomrule[1.5pt]
		\end{tabular}
	}
	\label{tab:model_arch}
\end{table}
\vspace{-10pt}

% \subsection{WenetSpeech Task}
% We first evaluate our proposed method on the WenetSpeech[20] Dataset, which a is multi-domain Mandarin corpus consisting of 10005 hours of high-quality labeled speech. Evaluation is performed in terms of word error rate (WER) on the Dev, Test Net and Test Meeting, described in [20].

\subsection{Main Results on 160k hours}
\label{sec:main_result}

\begin{table*}[!ht]
	\centering
	\caption{Following the scaling law \cite{DBLP:journals/corr/abs-2001-08361},  we compare model WERs on a fixed dataset (160k hours) across equal training steps (236k steps) or compute time (25.9 days). For the average WERs, we provide the mean values derived from two methods: the average obtained by summing up individual WERs (left) / the overall WER calculated by summing the insertion, deletion, and substitution errors for each item and then dividing by the total number of characters in the reference (right).}
	\scalebox{0.8}{
		\begin{tabular}{c|cc|c|cc}
			\toprule[1.5pt]
			(a) TestSet & (b) Dense-225M & (c) Dense-225M & (d) Dense-1B & (e) MoE-1B & (f) MoE-1B \\
                & 236k steps, 9.3 days & 657k steps, 25.9 days & 236k steps, 25.9 days & 236k steps, 16.8 days & 364k steps, 25.9 days \\
			\midrule
                speechio\_001 & 1.28 & 1.15 & 0.92 & 0.95 & 0.90  \\
                \midrule
                speechio\_002 & 3.51 & 3.30 & 3.03 & 3.08 & 2.94  \\
                \midrule
                speechio\_003 & 2.34 & 2.11 & 1.74 & 1.68 & 1.63  \\
                \midrule
                speechio\_004 & 2.05 & 1.96 & 1.79 & 1.87 & 1.93  \\
                \midrule
                speechio\_005 & 2.06 & 1.92 & 1.84 & 1.78 & 1.73  \\
                \midrule
                speechio\_006 & 7.24 & 6.69 & 6.34 & 6.35 & 6.34 \\
                \midrule
                speechio\_007 & 10.23 & 10.12 & 8.77 & 9.67 & 9.23 \\
                \midrule
                speechio\_008 & 7.34 & 6.29 & 5.78 & 6.13 & 5.59 \\
                \midrule
                speechio\_009 & 3.94 & 3.67 & 3.45 & 3.60 & 3.52 \\
                \midrule
                speechio\_010 & 4.76 & 4.68 & 4.37 & 4.55 & 4.49 \\
                \midrule
                speechio\_011 & 3.21 & 2.88 & 2.31 & 2.36 & 2.28 \\
                \midrule
                speechio\_012 & 3.39 & 3.22 & 2.91 & 3.01 & 2.97 \\
                \midrule
                speechio\_013 & 4.15 & 3.81 & 3.62 & 3.71 & 3.69 \\
                \midrule
                speechio\_014 & 5.01 & 4.45 & 3.87 & 4.06 & 3.83 \\
                \midrule
                speechio\_015 & 7.58 & 6.77 & 6.43 & 6.69 & 7.03 \\
                \midrule
                speechio\_016 & 5.15 & 4.46 & 3.95 & 4.02 & 3.82 \\
                \midrule
                speechio\_017 & 4.11 & 3.87 & 3.24 & 3.52 & 3.49 \\
                \midrule
                speechio\_018 & 2.69 & 2.57 & 2.38 & 2.56 & 2.44 \\
                \midrule
                speechio\_019 & 3.91 & 3.29 & 2.95 & 3.05 & 2.90 \\
                \midrule
                speechio\_020 & 3.05 & 2.97 & 2.33 & 2.51 & 2.47 \\
                \midrule
                speechio\_021 & 2.75 & 2.89 & 2.53 & 2.73 & 2.73 \\
                \midrule
                speechio\_022 & 5.55 & 5.15 & 4.50 & 4.86 & 4.52 \\
                \midrule
                speechio\_023 & 6.05 & 5.99 & 4.89 & 5.86 & 5.25 \\
                \midrule
                speechio\_024 & 5.61 & 5.19 & 4.61 & 4.76 & 4.78 \\
                \midrule
                speechio\_025 & 5.76 & 5.30 & 4.36 & 4.83 & 4.61 \\
                \midrule
                speechio\_026 & 4.37 & 4.01 & 3.90 & 4.02 & 3.84 \\
                \midrule
                average       & 4.50 / 3.79 & 4.18 / 3.55 & 3.72 / 3.16 & 3.93 / 3.36 & 3.80 / 3.22 \\
			\bottomrule[1.5pt]
		\end{tabular}
	}
	\label{tab:results_wer}
\end{table*}

In Table.\ref{tab:results_wer}, we compare the performance of the three models from Table.\ref{tab:model_arch} under different conditions (such as the same number of training steps or the same training time), with the results indicating:
\begin{itemize}
    \item[1)] At the same number of training steps (263k steps), comparing columns (b), (d), and (e) reveals that the WER of the MoE-1B model (3.93) is slightly worse than that of the Dense-1B model (3.72), but both significantly outperform the Dense-225M baseline (4.50).
    \item[2)] With the same training time (25.9 days), comparing columns (c), (d), and (f) shows that the WER of the MoE-1B model (3.80) is very close to that of the Dense-1B model (3.72), and both substantially surpass the Dense-225M model (4.18).
\end{itemize}

These results suggest that on a dataset of 160k hours, a larger number of parameters (from 225M to 1B) leads to better model performance. Moreover, when the number of parameters is the same, MoE models can achieve WER levels comparable to Dense models.

Furthermore, in Table.\ref{tab:results_rtf}, we compare the inference speeds of the three models, with the results showing:
\begin{itemize}
    \item[1)] Although the MoE-1B and Dense-1B have the same number of parameters, the former is 2.5 times faster than the latter.
    \item[2)] Even though the parameter count of MoE-1B is 4.7 times that of Dense-225M, the absolute difference in RTF between the two is only around 0.03 (for cpu) or 0.0004 (for gpu).
\end{itemize}

Overall, combining the WER and RTF results, we can confirm that \textbf{\textit{the MoE-1B model can achieve Dense-1B level accuracy with Dense-225M level inference cost}}.

% 同等rtf下acc更好
% 同等acc下rtf更好
% 做成一个图的形式表达上面两个点
% Scaling Vision with Sparse Mixture of Experts \cite{DBLP:conf/nips/RiquelmePMNJPKH21} 这篇文章附录的图表可以拿来借鉴

\begin{table}[!ht]
	\centering
	\caption{RTF benchmark. When testing with a CPU, we set the batch size to 1 and perform inference on an int8 quantized model using a single thread on an Intel(R) Core(TM) i5-8400 CPU @ 2.80GHz. For GPU-based evaluations, we set the batch size to 200 and perform inference on an FP16 model using a single NVIDIA 3090. Please note that we do not include GPU RTF for decoder rescoring since the inference time for this process is dominated by the CTC prefix beam search running on the CPU, and therefore, it cannot objectively reflect the inference time on the GPU.}
	\scalebox{0.85}{
		\begin{tabular}{c|cc}
			\toprule[1.5pt]
			(a) Model & (b) ctc greedy decoding & (c) decoder rescoring \\
                \midrule
                Dense-225M & 0.1088 (cpu) / 0.0012 (gpu) & 0.1524 (cpu) \\
                Dense-1B & 0.3155 (cpu) / 0.0028 (gpu) & 0.4515 (cpu) \\
                MoE-1B & 0.1299 (cpu) / 0.0016 (gpu) & 0.1826 (cpu) \\
			\bottomrule[1.5pt]
		\end{tabular}
	}
	\label{tab:results_rtf}
\end{table}

\subsection{Streaming Capability}

Empirically, training a large model to accommodate both streaming and non-streaming modes simultaneously could potentially compromise performance. In response, this paper introduces a two-stage training pipeline. Initially, we train a non-streaming base model (such as MoE-1B and Dense-225M that is described in Section \ref{sec:train_detail} and Table \ref{tab:model_arch}), which then serves as the foundation for initializing the proposed U2++-MoE-1B model (and also U2++-Dense-225M, U2++-Dense-1B). The MoE-1B model shares an identical architecture with the U2++-MoE-1B model, with the only distinction lying in their approach to chunk masking. While the MoE-1B employs a full chunk strategy, the U2++-MoE-1B adopts a dynamic chunk method as detailed in section \ref{sec:method}. This approach stabilizes the training process for a unified system capable of handling both streaming and non-streaming functionalities.

In Table \ref{tab:results_stream}, by comparing three different streaming models, we can draw the same conclusion as in the non-streaming models (section \ref{sec:main_result}), which is that our proposed MoE model significantly outperforms the Dense counterpart in terms of WER while maintaining a similar RTF. Please note that the WER for the U2++-Dense-1B model is not included. This is due to the frequent occurrence of gradient explosions during the training process, which, despite the initialization with a non-streaming Dense-1B model, made the training unsustainable.

\begin{table}[!ht]
	\centering
	\caption{Averaged streaming results on SpeechIO test sets: WER Measured with a 640ms chunk size and RTF calculated using the same hardware (cpu) and methodology (decoder rescoring) as in Table \ref{tab:results_rtf}. All models were initialized from their respective non-streaming baselines and subsequently trained for a total of 160k steps.}
	\scalebox{1.0}{
		\begin{tabular}{c|cc}
			\toprule[1.5pt]
			(a) Model & (b) WER & (c) RTF \\
                \midrule
                U2++-Dense-225M & 6.24 & 0.1937 \\
                U2++-Dense-1B & N/A & 0.6015 \\
                U2++-MoE-1B & 4.83 & 0.2436 \\
			\bottomrule[1.5pt]
		\end{tabular}
	}
	\label{tab:results_stream}
\end{table}

\iffalse
\subsection{Scaling Law}
\begin{table}[!ht]
	\centering
	\caption{wenetspeech results.}
	\scalebox{0.85}{
		\begin{tabular}{c|ccc}
			\toprule[1.5pt]
			(a) Model & (b) dev & (c) test\_net & (d) test\_meeting \\
                \midrule
                Dense-116M & 5.90 & 8.96 & 11.99 \\
                Dense-225M & 4.42 & 6.50 & 9.02 \\
                Dense-1B &  &  & \\
                MoE-1B & 4.32 & 6.3 & 8.49 \\
			\bottomrule[1.5pt]
		\end{tabular}
	}
	\label{tab:results_wenetspeech}
\end{table}
\fi

% \subsection{Further analyze on expert activation}

% We further inspect whether expert sparsity exists on the encoder and the decoder of the model. Figure 7 demonstrates the expert sparsity level on the encoder and decoder on all three tasks. We find that the encoder activation is mostly dense, that most of the experts are activated at all times. The decoder activation is extremely sparse (about 75\%).

\section{Conclusion}

% We provide optimization strategies for efficient MoE deployment, reducing inference costs with minimal impact on model quality.

The proposed U2++ MoE provides a clean setup and little task-specific design. Through the straightforward substitution of all FFN layers in the baseline model with MoE FFNs, coupled with the adoption of the U2++ training framework, we attain notable enhancements in WER alongside streaming recognition capabilities, all without a considerable increase in RTF.

\section{Acknowledgements}
We thank Wenpeng Li and Jianwei Niu for their feedbacks on this work.

\vfill\pagebreak

% References should be produced using the bibtex program from suitable
% BiBTeX files (here: strings, refs, manuals). The IEEEbib.bst bibliography
% style file from IEEE produces unsorted bibliography list.
% -------------------------------------------------------------------------
\bibliographystyle{IEEEbib}
\bibliography{strings,refs}

\end{document}